\documentclass[journal]{IEEEtran}
\ifCLASSINFOpdf
\else
\fi
\usepackage{epsfig}
\usepackage{graphicx}
\usepackage{amsmath}
\usepackage{amssymb}
\usepackage{algorithm}
\usepackage{algorithmic}
\usepackage{multirow}
\usepackage{graphicx}
\usepackage{subfigure}
\usepackage{caption}
\usepackage{soul,color}
\usepackage{color}
\usepackage{amsfonts,amssymb}
\usepackage{booktabs}
\usepackage{CJK}
\usepackage{multirow}
\usepackage{makecell}
\usepackage{cite}
\usepackage[numbers,sort&compress]{natbib}

\begin{document}
\title{SegGPT Meets Co-Saliency Scene}

\author{Yi~Liu,
Shoukun~Xu,
Dingwen~Zhang$\dag$
and~Jungong~Han
%and~Long~Wang
%and~Ling~Shao,~Senior~Member~IEEE% <-this % stops a space
\thanks{Yi Liu and Shoukun Xu are with School of Computer Science and Artificial Intelligence, Aliyun School of Big Data, and School of Software, Changzhou University, Changzhou, Jiangsu, 213000, China. Email: liuyi0089@gmail.com, jpuxsk@163.com.}
%\thanks{Nian Liu is with Mohamed bin Zayed University of Artificial Intelligence, Abu Dhabi, UAE. Email: liunian228@gmail.com.}
\thanks{Dingwen Zhang is with the Hefei Comprehensive National Science Center, Institute of Artificial Intelligence, Hefei 230026, China, and School of Automation, Northwestern Polytechnical University, Xi'an, Shannxi, 710129, China. Email: zhangdingwen2006yyy@gmail.com.}
\thanks{Jungong Han is with Department of Computer Science, The University of Sheffield, U.K. Email: jungonghan77@gmail.com.}
\thanks{$^\dag$: Corresponding author.}
%\thanks{Long Wang is with Center for Systems and Control, College of Engineering, Peking University, Beijing
%100871, China. Email: longwang@pku.edu.cn.}
% <-this % stops a space
%\thanks{Manuscript received XX XX, 2016.}
}

% ^------------^------------^----Do not want these spaces!

\markboth{}%
{Shell \MakeLowercase{\textit{et al.}}: Bare Demo of IEEEtran.cls for IEEE Journals}
% *** name in the headers of peer review papers. ***



\maketitle

\begin{abstract}
Co-salient object detection targets at detecting co-existed salient objects among a group of images. Recently, a generalist model for segmenting everything in context, called SegGPT, is gaining public attention. In view of its breakthrough for segmentation, we can hardly wait to probe into its contribution to the task of co-salient object detection. In this report, we first design a framework to enable SegGPT for the problem of co-salient object detection. Proceed to the next step, we evaluate the performance of SegGPT on the problem of co-salient object detection on three available datasets. We achieve a finding that co-saliency scenes challenges SegGPT due to context discrepancy within a group of co-saliency images.
\end{abstract}

\begin{IEEEkeywords}
Co-Salient object detection, SegGPT, Context
\end{IEEEkeywords}

% \fi
\IEEEpeerreviewmaketitle

\section{Introduction}
\label{sec:Introduction}

\IEEEPARstart{C}{o-salient} object detection aims to detect to common salient objects among a group of input images. Unlike salient object detection, which is to detect the most attractive objects by mimicking human eyes \cite{borji2015salient, wang2021salient}, co-salient object detection focuses on detecting salient and co-existed objects among all the input images.

Recently, there has emerged a powerful model called SegGPT \cite{wang2023seggpt} for segmentation. SegGPT \cite{wang2023seggpt} is capable of segmenting everything  in context. Inspired by its breakthrough for the computer vision community, we have a great mind to study its contribution to the problem of co-salient object detection.

In this report, a framework is first designed to enable SegGPT \cite{wang2023seggpt} in the task of co-salient object detection. On top of evaluation, a discussion is presented for the involvement of SegGPT \cite{wang2023seggpt} for co-saliency scenes.

\section{Methodology}

\begin{figure}[t]
\centering
 \includegraphics[width=0.98\linewidth]{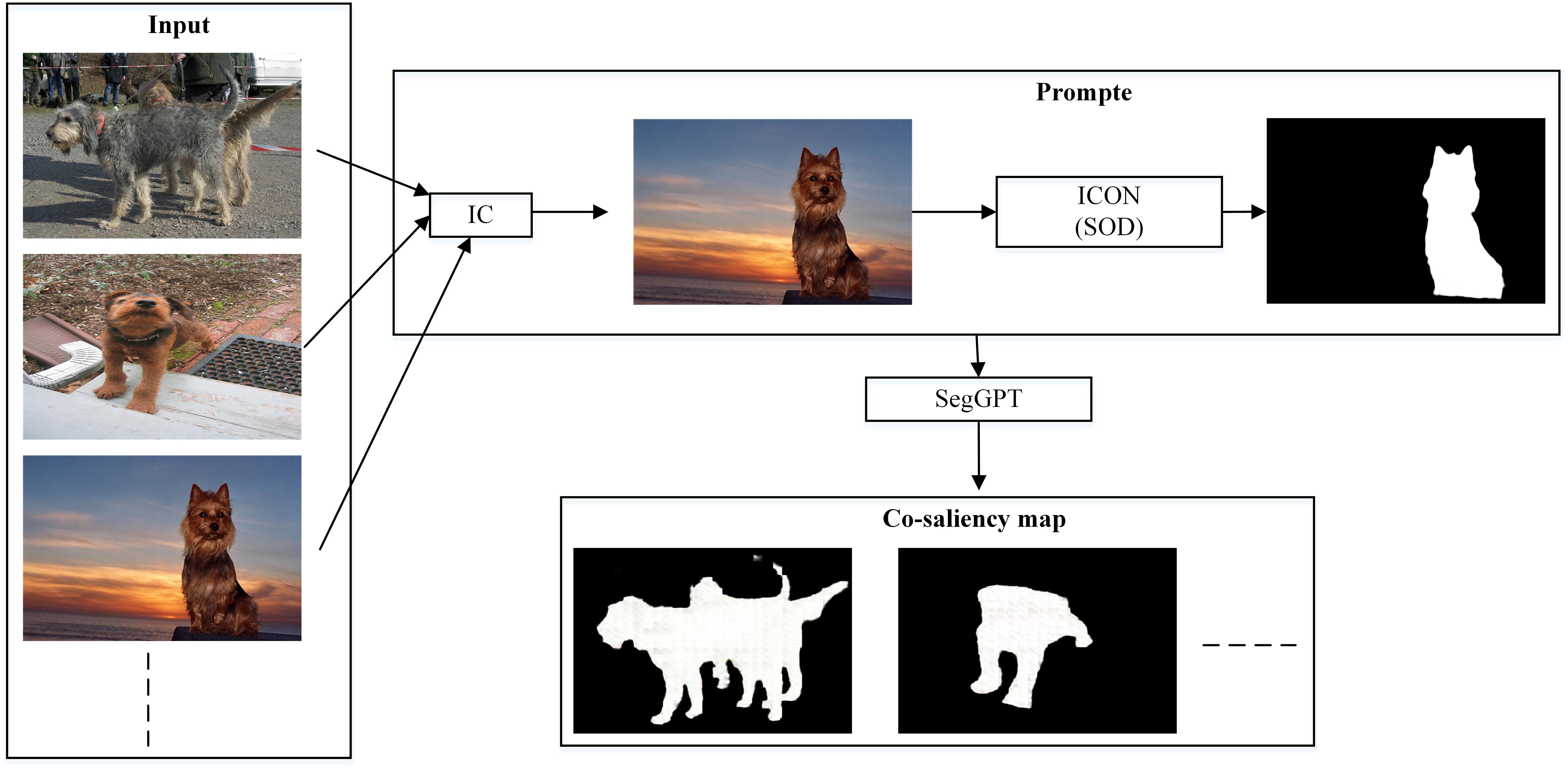}
%\vspace{-1.2em}
%\captionsetup{font={footnotesize}}
\caption{Overview of the framework. The input images are first input in the IC \cite{feng2022ic9600} algorithm to select the simple-scene image, which is followed by a salient object detector, \emph{i.e.}, ICON \cite{zhuge2022salient}, to generate the prompt. On top of that, the prompt image and saliency map are fed in SegGPT \cite{wang2023seggpt} to predict the co-salient objects in all target images to infer the co-salient maps.}
\label{fig:framework}
%\vspace*{-15pt}
\end{figure}

The overview of the framework is shown in Fig. \ref{fig:framework}. To generate a high-quality prompt, we employ a salient object detector to infer the salient object of a simple-scene image from the group of images. To select the simple-scene image, we adopt the IC algorithm \cite{feng2022ic9600} to compute the complexity of images within the group. The lowest-complexity image is chosen as the prompt image, which is detected by ICON \cite{zhuge2022salient}, which is a salient object detector, to generate the prompt segmentation. The prompt image and segmentation are fed in SegGPT \cite{wang2023seggpt} to infer co-salient maps for the group of images.

\begin{figure}[t]
\centering
 \includegraphics[width=0.98\linewidth]{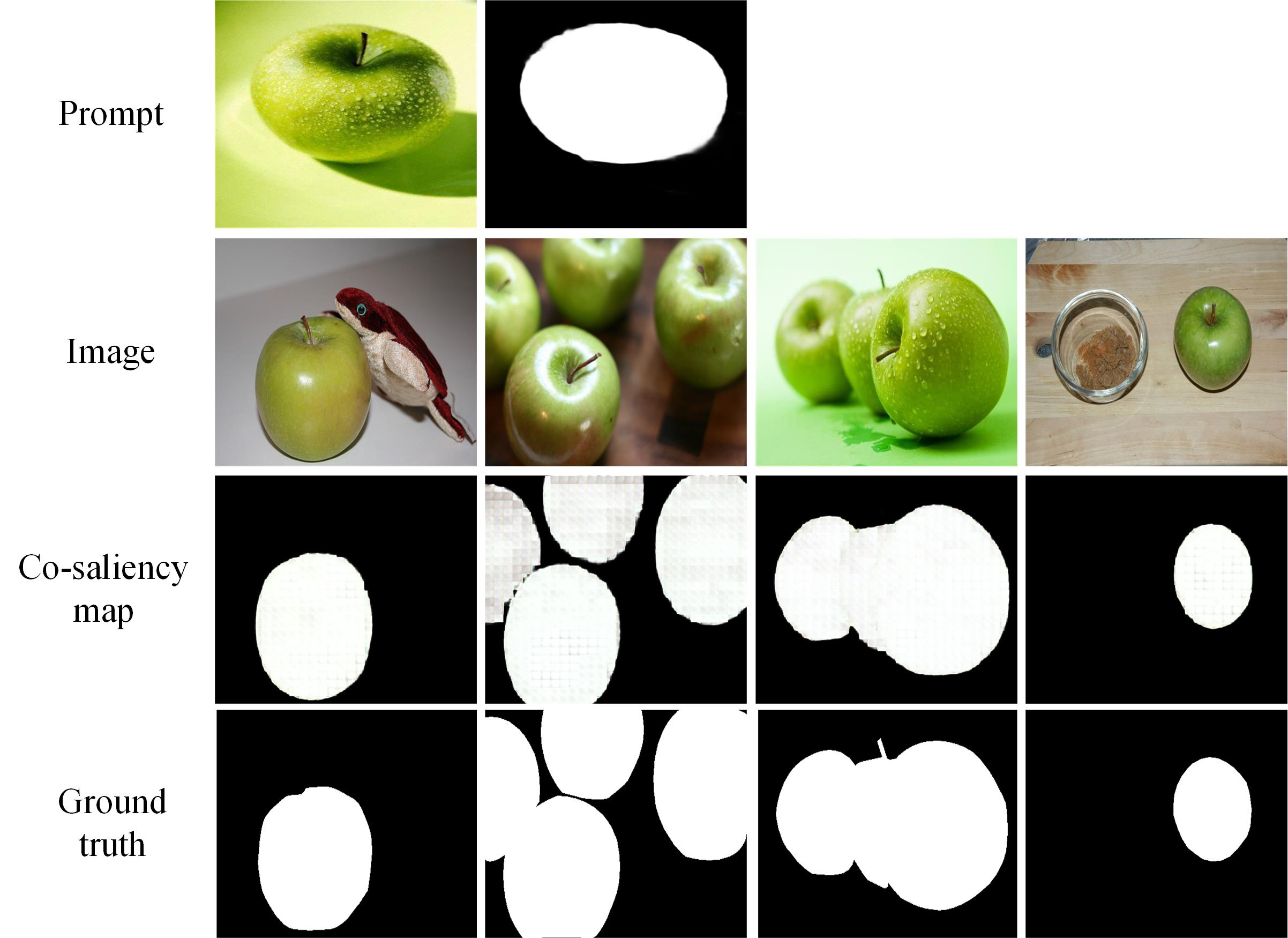}
%\vspace{-1.2em}
%\captionsetup{font={footnotesize}}
\caption{Visual results of the proposed framework.}
\label{fig:visual-result}
%\vspace*{-15pt}
\end{figure}

Fig. \ref{fig:visual-result} displays the detection results of the proposed framework. We can find that SegGPT \cite{wang2023seggpt} can well segment all co-salient objects based on the prompt image and map.

\section{Experiment and Analysis}
\label{sec:Experiment}
In this section, we conduct experiments to investigate the proposed framework for co-saliency detection.
\begin{table*}[htbp]
	\centering
%    \Large
%    \captionsetup{font={footnotesize}}
	\caption{$F_\beta^{max}$, MAE, $S_m$, and $E_m^{max}$ values of different methods. The best method is marked by \textbf{bold}. The symbols $\uparrow$/$\downarrow$ mean that a higher/lower score is better. The difference between our proposed framework and the best cutting-edge method is marked by {\color{blue}blue}.}
%\vspace{-1.2em}
	\label{tab:metric}
	\scalebox{0.86}{
			\begin{tabular}{c|c|c|c|c|c|c|c|c|c|c|c|c}
				\hline
				\multirow{4}{*}{}&
				\multicolumn{4}{c}{\textbf{CoSOD3k \cite{fan2021re}}}&
				\multicolumn{4}{|c}{\textbf{CoCA \cite{zhang2020gradient}}}&\multicolumn{4}{|c}{\textbf{CoSal2015 \cite{zhang2016detection}}}\cr\cline{2-13}
				&$F_\beta^{max}\uparrow$&MAE$\downarrow$&$S_m\uparrow$&$E_m^{max}\uparrow$&$F_\beta^{max}\uparrow$&MAE$\downarrow$&$S_m\uparrow$&$E_m^{max}\uparrow$&$F_\beta^{max}\uparrow$&MAE$\downarrow$&$S_m\uparrow$&$E_m^{max}\uparrow$\cr
				\hline
                \hline
\bf{CSMG$_{CVPR 2019}$ \cite{zhang2019co}}&0.7297&0.1480&0.7272&0.8208&0.4988&0.1273&0.6276&0.7324&0.7869&0.1309&0.7757&0.8436\cr
				\hline
\bf{GICD$_{ECCV 2020}$ \cite{zhang2020gradient}}&0.7698&0.0794&0.7967&0.8478&0.5126&0.1260&0.6579&0.7149&0.8441&0.0707&0.8437&0.8869\cr
				\hline
\bf{ICNet$_{NeurIPS 2020}$ \cite{jin2020icnet}}&0.7623&0.0891&0.7942&0.8450&0.5133&0.1470&0.6541&0.7042&0.8583&0.0579&0.8571&0.9011\cr
				\hline
\bf{GCoNet$_{CVPR 2021}$ \cite{fan2021group}}&0.7771&0.0712&0.8018&0.8601&0.5438&0.1050&0.6730&0.7598&0.8471&0.0681&0.8453&0.8879\cr
				\hline
\bf{CADC$_{ICCV 2021}$ \cite{zhang2021summarize}}&0.7781&0.0875&0.8150&0.8543&0.5487&0.1330&0.6800&0.7443&0.8645&0.0641&0.8666&0.9063\cr
				\hline
\bf{DCFM$_{CVPR 2022}$ \cite{yu2022democracy}}&0.8045&0.0674&0.8094&0.8742&0.5981&\bf{0.0845}&0.7101&0.7826&0.8559&0.0672&0.8380&0.8929\cr
				\hline
\bf{DMT$_{CVPR 2023}$ \cite{li2023discriminative}}&\bf{0.8353}&\bf{0.0633}&\bf{0.8514}&\bf{0.8950}&\bf{0.6190}&0.1084&\bf{0.7246}&\bf{0.8001}&\bf{0.9052}&\bf{0.0454}&\bf{0.8974}&\bf{0.9362}\cr
				\hline
\bf{OURS}&0.7560&0.0804&0.7997&0.8364&0.4880&0.0989&0.6474&0.6855&0.7812&0.0746&0.8216&0.8539\cr
				\hline
\bf{Difference}&{\color{blue}-7.93\%}&{\color{blue}+6.33\%}&{\color{blue}-5.17\%}&{\color{blue}-5.86\%}&{\color{blue}-13.10\%}&{\color{blue}+1.44\%}&{\color{blue}-7.72\%}&{\color{blue}-11.46\%}&{\color{blue}-12.40\%}&{\color{blue}+2.92\%}&{\color{blue}-7.58\%}&{\color{blue}-8.23\%}\cr
				\hline
			\end{tabular}}
\end{table*}
\begin{figure*}[t]
\centering
 \includegraphics[width=0.98\linewidth]{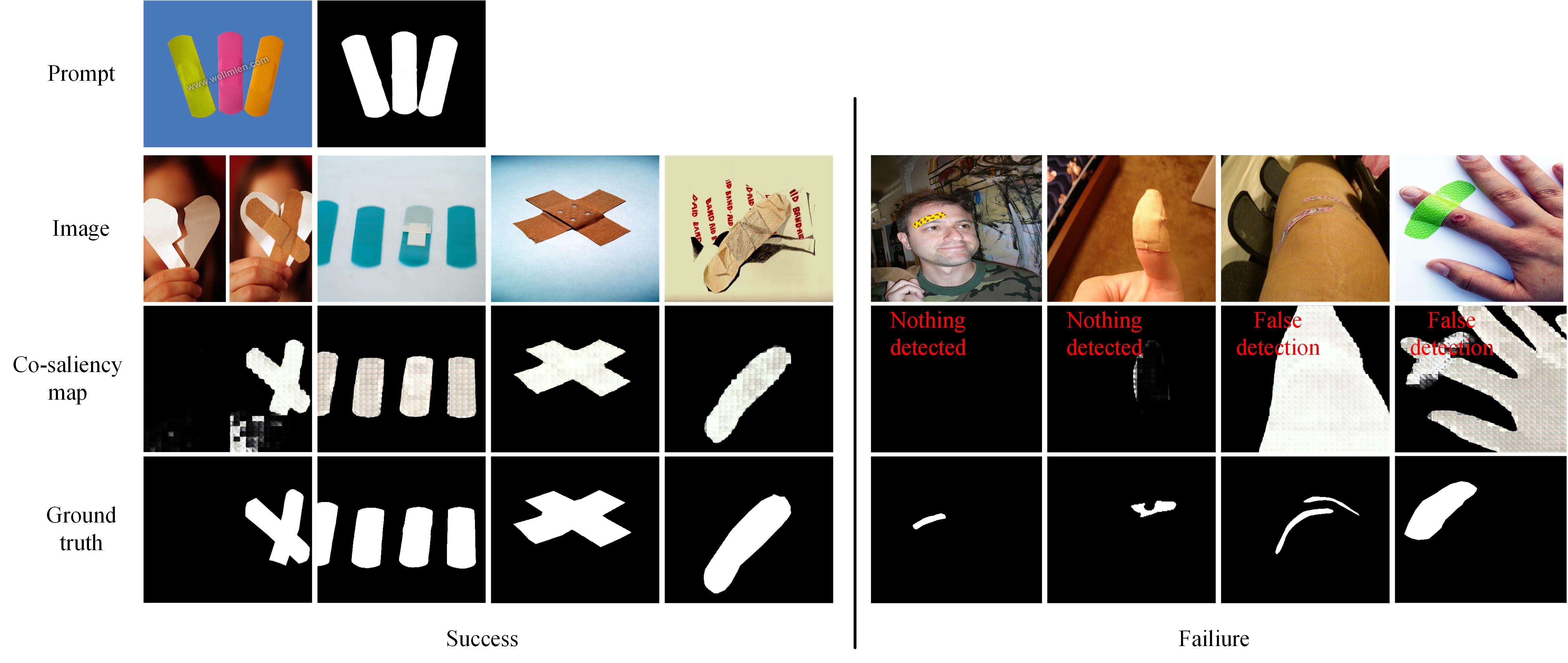}
%\vspace{-1.2em}
%\captionsetup{font={footnotesize}}
\caption{Success and failure cases of the proposed framework. For success cases, since these images share similar context with the prompt image, their co-existed salient objects are identified by SegGPT \cite{wang2023seggpt}. By contrast, those images in failure cases pose complex context and sharply different context with the prompt image, because of which SegGPT \cite{wang2023seggpt} will detect nothing or generate false detections.}
\label{fig:visual-result1}
%\vspace*{-15pt}
\end{figure*}
\subsection{Evaluation protocols}

\textbf{Dataset.} To verify the effectiveness of the proposed framework, we conduct experiments on three public co-salient object detection datasets, including CoSOD3k \cite{fan2021re}, CoCA \cite{zhang2020gradient}, and CoSal2015 \cite{zhang2016detection}. \textbf{CoSOD3k \cite{fan2021re}} and \textbf{CoSal2015 \cite{zhang2016detection}} collects 50 groups with 2015 images and 160 groups with 3316 images, respectively.
\textbf{CoCA \cite{zhang2020gradient}} is the most challenging ataset and contain 1295 images of 80 groups.

\textbf{Evaluation metrics.} We select four evaluation metrics to evaluate the performance of different methods, including maximum F-measure ($F_\beta^{max}$) \cite {achanta2009frequency}, MAE \cite{cheng2013efficient}, structure-measure ($S_m$) \cite{fan2017structure}, and enhanced-alignment measure ($E_m$) \cite{fan2018enhanced}.

\textbf{State-of-the-art methods.} We compare our proposed framework with 7 state-of-the-art methods, including CSMG \cite{zhang2019co}, GICD \cite{zhang2020gradient}, ICNet \cite{jin2020icnet}, GCoNet \cite{fan2021group}, CADC \cite{zhang2021summarize}, DCFM \cite{yu2022democracy}, and DMT \cite{li2023discriminative}.
\subsection{Evaluation and discussion}

Table \ref{tab:metric} lists performance of different methods. It can be found that our proposed framework is inferior to the cutting-edge method with a significant gap. Some success and failure cases are depicted in Fig. \ref{fig:visual-result1}, which indicates two findings:

i) The images in success cases share highly-similar context with the prompt image, their co-existed salient objects can be well identified.

ii) The images in failure cases, although sharing the same salient objects, pose different contexts due to the introduction of complicated scene, \emph{e.g.}, right-hand four images. Thus their co-existed salient objects will be hardly detected due to the complex context.

To sum up, we can come to a conclusion. The in-context capability of SegGPT \cite{wang2023seggpt} is able to solve segmentation in those images sharing similar context with the prompt image, but fail at segmentation in those images posing different contexts besides the prompt context.

\section{Conclusions}
\label{sec:Conclusion}

In this paper, we have conducted an empirical study of SegGPT on the problem of co-salient object detection. First, we design a framework to introduce SegGPT for the task setting of co-salient object detection. Secondly, we evaluate the performance and provide an investigation of SegGPT on co-salient object detection. We expect this paper will present some inspiration for the researchers on the task of co-salient object detection, and help them put up new ideas for this field.

%\section*{Acknowledgment}
%
%This work is supported by the National Natural Science Foundation of China under Grant No. 62001341 and the National Natural Science Foundation of Jiangsu Province under Grant No. BK20221379.
\bibliographystyle{IEEEtran}
\bibliography{ref}

\end{document}